\documentclass[]{article}

\usepackage[utf8]{inputenc}
\usepackage{graphicx}
\usepackage{vhistory}
\usepackage{hyperref} 
\usepackage{natbib}
\usepackage{amsmath} 
\usepackage[ruled,vlined]{algorithm2e}
\usepackage{dsfont}
\usepackage{bbm}

\usepackage{authblk}
\newcommand{\one}{ \mathds{1} }
\newcommand{\real}{\rm I\!R}
\newcommand{\ELBO}{\mathcal{L}}
\newcommand{\lr}{\ell}
\newcommand{\batchsize}{B}
\newcommand{\perm}{\mathcal{P}}

\newcommand{\PT}{P^{\theta}}
\newcommand{\rel}{r^{\theta}}
\providecommand{\keywords}[1]
{
  \small	
  \textbf{\textit{Keywords---}} #1
}

\begin{document}

\newtheorem{strategy}{Recommendation Strategy}
\title{Dynamic Slate Recommendation with Gated Recurrent Units and Thompson Sampling}


\author[1]{Simen Eide}
\affil[1]{Department of Mathematics, University of Oslo and FINN.no}
\author[2]{David S. Leslie}
\affil[2]{Department of Mathematics and Statistics, Lancaster University}
\author[3]{Arnoldo Frigessi}
\affil[3]{Oslo Centre for Biostatistics and Epidemiology, University of Oslo}




\date{}

\maketitle

\begin{abstract}

We consider the problem of recommending relevant content to users of an internet platform in the form of lists of items, called slates.
We introduce a variational Bayesian Recurrent Neural Net recommender system that acts on time series of interactions between the internet platform and the user, and which scales to real world industrial situations.
The recommender system is tested both online on real users, and on an offline dataset collected from a Norwegian web-based marketplace, FINN.no, that is made public for research.
This is one of the first publicly available datasets which includes all the slates that are presented to users as well as which items (if any) in the slates were clicked on.
Such a data set allows us to move beyond the common assumption that implicitly assumes that users are considering all possible items at each interaction. 
Instead we build our likelihood using the items that are actually in the slate, and evaluate the strengths and weaknesses of both approaches theoretically and in experiments.
We also introduce a hierarchical prior for the item parameters based on group memberships. 
Both item parameters and user preferences are learned probabilistically.
Furthermore, we combine our model with bandit strategies to ensure learning, 
and introduce `in-slate Thompson Sampling' which makes use of the slates to maximise explorative opportunities. We show experimentally that explorative recommender strategies perform on par or above their greedy counterparts. Even without making use of exploration to learn more effectively, click rates increase simply because of improved diversity in the recommended slates.

\end{abstract}

\section{Introduction}

Online services and marketplaces often contain millions or billions of potential items that a user can consume, making the task of finding relevant items challenging for the user.
Platforms often present their content as lists of items, usually called feeds or slates \citep{Ie2019a, Bello}.
A search slate is generated from an explicit query from the user, whereas a recommendation slate is usually based on the user's previous interactions on the site. 
The user can decide how far to scroll the slate and can click on any item that interests her, providing the platform with an implicit feedback signal of the user's preferences \citep{Hu2008}.
In this article we consider methods to construct recommendation slates to users on the basis of their history of interaction with the system.

We know that a user's preference is dynamic, and will deploy methods of sequential recommendations.
A sequential recommender system models the interest of users, which may change over time \citep{Ying2018a}.
This is in contrast to a stationary recommender system that assumes that the user's preferences do not change over time.
In many situations it is important to allow changing user interest over time.
For example, a user will typically no longer be interested in a computer after the point of purchase.

The aim of the recommender system is to recommend items that the user wants to click on, by repeatedly interacting with the user and presenting slates consisting of items that either the system expects the user to click on, or that the system wishes to learn the users' interest about.
We take a contextual multi-armed bandit approach \citep{Lattimore2019} to the problem where the recommender system is the agent and the users are the environment.
We develop a Bayesian gated recurrent neural net model that is trained on all users' past behavior and estimates the click probability of each item in a given slate based on the click history of the current user.
Building on this model, we propose a bandit-based recommender strategy that suggests relevant items to the users.

\paragraph{Slate model}
In each interaction the user considers the set of items in the slate and decides which item to click on, if any.
We can therefore model the click probability as a categorical distribution over all items in the slate, including an additional `no-click item'. 
We refer to the resulting likelihood as the `slate likelihood'.
Although slate recommendation is a very common problem in the industry, to our knowledge, there is no publicly available dataset that presents all items that were presented to the users in the slates including those that were not clicked, which we term `exposure data'.
In this article we produce such a dataset and suggest it as benchmark for future research.
In the absence of exposure data, a common model assumption in recommender systems is the `all-item likelihood', sometimes called uniform negative sampling \citep{Covington2016}, which implicitly assumes that the user considers all items on the platform in each interaction instead of only those presented in the slate; other approaches are presented by \citet{Tran2019, Rendle2009} and are discussed in Section \ref{sec:relatedwork}.
In Section \ref{sec:slatevsall-item} we clarify an important aspect, namely that the all-item likelihood implicitly models the joint distribution of clicks and exposures, in the form of
the slate likelihood multiplied by a specific generative exposure model that is proportional to the user's preferences. 
Contrary to intuition, we show both theoretically and empirically that in some important cases the all-item likelihood may actually improve the performance of the recommender system compared to the more ambitious slate likelihood.

\paragraph{Bayesian bandit model}
Although recommender systems have large volumes of training data, the amount of data for each user and item may be limited. Inactive users and new or unpopular items have very few interactions with the platform, giving rise to the common cold-start problem in recommender systems.
The cold start problem motivates a Bayesian approach to recommendation systems for two reasons.
First, a Bayesian approach allows for the introduction of hierarchical priors that share information probabilistically between related parameters.
This is particularly useful for item parameters, because we often know that groups of items share certain properties (e.g. items belonging to the user-defined product category `sofa' are likely to have similar model parameters).
Hierarchical priors have previously been used in stationary models \citep{Gopalan2013}, but has to our knowledge not been applied to sequential recommender systems.
Second, the posterior distribution of click probabilities allows for the use of Bayesian exploration techniques, such as Thompson sampling, to explore the users' preferences when recommending items.
Importantly, by quantifying the uncertainty of user-specific parameters, we can present a more diverse slate to the user \citep[as in][]{Edwards2018,Edwards2019}.
This can lead to an increase in click rates, as it is more likely that the user finds a relevant item in a diverse slate, while also improving understanding of the preferences of the user.
This article uses both of these Bayesian techniques to improve the recommender system:
we show that using hierarchical item priors improves both offline and online metrics, and that explorative recommender strategies, such as Thompson sampling, can increase click rates through improved diversity with no additional exploration cost. 

\paragraph{Slate Dataset}
To address the lack of open slate datasets with exposure data, we release a dataset from the online marketplace \href{https://finn.no}{FINN.no} for research purposes. 
It includes both search and recommendation interactions between users and the platform over a 30 day period.
The dataset has logged both exposures and clicks, including interactions where the user did not click on any of the items in the slate.
This unique dataset is presented in more detail in Section \ref{sec:dataset}.
The data, as well as the code used in this article, are available at \href{https://github.com/finn-no/recsys-slates-dataset}{https://github.com/finn-no/recsys-slates-dataset}.

\paragraph{Contributions and structure}
The main contributions of this article are summarised here:
(1) We develop a sequential slate likelihood model and compare it to the more commonly used all-item likelihood model. 
We show that the all-item likelihood implicitly learns the users' preferences from the slates the platform presents to the user, as well as from the items the user clicks.
This allows a system using the all-item likelihood to outperform one using the slate likelihood when the platform presents informed slates to the user, which is typically the case when user interactions are dominated by searches.
On the other hand, the slate likelihood outperforms the all-item likelihood when there are less search slates in the dataset.
(2) We build a variational Bayes recommender system with hierarchical item priors and a gated recurrent neural net to model the user dynamics that scales to an industrial level dataset.
This allows us to introduce an explorative `in-slate Thompson sampling' recommender strategy that increases click rates, not through improved learning, but by increasing the diversity in the slates.
(3) We publish one of the first slate datasets for research purposes, which come from the online marketplace \href{https://finn.no}{FINN.no}.

The article is organized as follows. Section \ref{sec:relatedwork} reviews the relevant literature. 
We define the sequential slate recommendation problem in Section \ref{sec:problem}.
Section \ref{sec:model} describes the slate likelihood model with its properties and proposes variational inference.
Section \ref{sec:strategy} discusses different recommendation strategies.
In Section \ref{sec:dataset} we present the released slate dataset.
Section \ref{sec:experiments} is dedicated to one offline study and two online studies examining the effects of different model choices and recommendation strategies, and we conclude the article in Section \ref{sec:conclusion}.

\subsection{Related work} \label{sec:relatedwork}
This article builds upon many research areas, in particular sequential recommender systems, exposure assumptions in recommender systems, bandits and Bayesian recommender systems.
We also review the closest comparable public datasets to the one we publish in this paper.

\paragraph{Exposure modeling in recommender systems}
The most common approach to model exposures is to assume that the user evaluates all possible items (`all-item likelihood', defined in (\ref{eq:likelihood_all-item})), as used in e.g. \cite{Covington2016}.
There are also other approaches in the literature. There are multiple papers using different variations of ranking likelihoods \citep[e.g.][]{Rendle2009,Tran2019, Hsieh2017}, which are similar to the all-item likelihood, with non-clicked items weighted down in the likelihood.
For example, \cite{Rendle2009} presents a ranking likelihood called Bayesian Personalized Ranking (BPR) that gives clicked items higher relevance scores than those that are not clicked.
However, the non-clicked items do not originate from the actual exposures considered by the user, but from the full item universe.
The closest approach to ours in this respect is \cite{Ie2019a}, which describes a reinforcement learning recommender system with a logit click likelihood over exposed items in a slate, similar to the our slate likelihood (see (\ref{eq:click-softmax})).
However, \cite{Ie2019a} does not model the user preferences over time as we do in the current article. 
Instead the user is characterized by a static set of features at each interaction.
Furthermore, their experiment is evaluated using long term reward from the user, whereas we focus on click rates.
Finally, the present article uses Bayesian deep learning in inference and probabilistic recommender strategies, whereas \cite{Ie2019a} uses a frequentist maximum likelihood estimate.

\paragraph{Causal recommender systems}
While the focus of the present article is not causal inference, computing the probabilities that a user will click on an item which she has not yet been exposed to is counterfactual. 
We compute these probabilities in order to recommend the most relevant items.
\cite{Liang2016} developed a causal inference approach to recommender systems using a classic matrix factorization model.
Like in the present article, they propose a model for the exposure process, and perform Bayesian inference.
However, the model in \cite{Liang2016} is different because the user factors do not follow a stochastic process in time.
It also computes the maximum a posteriori point estimates, and does not consider the full posterior distribution which we exploit in this article to recommend items, using variational inference and Thompson sampling.
Further, \cite{Liang2016} assume that only one item at a time is recommended, and do not model that the user is interacting with the system over time.

\sloppy
\paragraph{Sequential Recommender Systems}
Recently, \cite{hidasi2016b} presented the gru4rec model, which is one of the first models that use Gated Recurrent Neural Nets to describe the user profile in recommender systems.
They show that gru4rec outperforms other methods typically used in recommender systems such as item-KNN \citep{linden2003}.
\cite{hidasi2016b} consider different likelihood functions including the ranking loss BPR \citep{Rendle2009}, which assumes that all items are ranked lower than the clicked item, similarly as in the all-item likelihood.
In addition, inference in \cite{hidasi2016b} is not Bayesian, nor do they consider any recommendation strategies that perform exploration.

\paragraph{Recommender systems as a bandit problem.}
Bandit type algorithms have recently become more popular in recommender systems.
\cite{Li2016} develops a collaborative filtering bandit using the Yahoo-news dataset (including exposure and click/noclick for news recommended at random).
However, that paper does not consider a dynamic user profile, nor slates.
\cite{Guo} test both Thompson sampling and Upper Confidence Bound strategies on both simulated and real data in a display advertising context.
The problem differs from the present paper in that \cite{Guo} only considers click rates when single items are presented, whereas the present article considers multiple items presented in slates.
\cite{Guo} used the dropout trick \citep{Gal2016}, but only in the last layer of the neural network, to obtain approximate posterior samples. 
Instead, the present article is using full scale Variational Inference.
\cite{Guo} also used the `wide and deep' model architecture as in \cite{Cheng2016}, where the users are characterized by a combination of covariates and some user-specific parameters, whereas the present article is using an autoregressive user dynamics model.
\cite{Guo} finds, as we also do in this article, that exploration does not have a large cost when running in a large-scale online environment.

\paragraph{Bayesian Recommender systems}
There are several Bayesian Matrix Factorization models in the literature \citep{Salakhutdinov2008, Li2019, Zhang2007, Ngaffo2020}.
Common for these is that they do not consider varying user profiles, they do not consider bandit effects and most consider ratings feedback instead of the implicit click.
\cite{Salakhutdinov2008} build a hierarchical prior but use only a global prior common to all items, and do not utilize the fact that some items share attributes. 
Interestingly, they get higher prediction accuracy compared to the MAP estimate, attributing it to reduced overfitting when including the prior.
\cite{Li2019} take the hierarchical prior idea further and infer a hidden hierarchical structure without the knowledge of any attributes that group the items, showing that this can outperform a model without groups.

\paragraph{Recommender datasets including exposures}
To our knowledge there does not exist a recommender system dataset that records all slate interactions over a long time period, recording both all exposures and all clicks.
The most similar dataset we could find relates to display of advertisements \citep{criteo}, which  contains both exposures and clicks.
However, the Criteo dataset only includes single item recommendations and not slates.
Parallel to our work, \cite{Rekabsaz2021} released an information retrieval dataset that includes click log entries from a health website.
They include the top 20 retrieved documents from each slate, but do not identify how far the user scrolled.
It is also focusing on queries, and the average number of slates shown to the user is 3.3, which is less than typically seen in recommendation applications and datasets.

\vskip 3 mm
\noindent
The present article builds on this large literature, combining some ideas and solutions with several new suggestions related to the sequential slate dynamics, the gated recurrent neural network, the group hierarchical models for item parameters, the fully Bayesian variational inference, and probabilistic recommendation strategies, fully implemented in a way which scales to industrial settings. Our approach is presented with experimental results on a new dataset which will help in future benchmarking of recommender systems.

\section{Problem formulation and notation} \label{sec:problem}
Let $I = \{i\}_{i=0}^{|I|}$ be a set of $|I|$ items and $U = \{u\}_{u=1}^{|U|}$ a set of $|U|$ users.
The zeroth item $i=0$ is a special item representing the action of no click.
Each item $i$ belongs to a group $g(i) \in G$, where the number of groups is much smaller than the number of items ($|G| \ll |I|$).
Each group is typically a category of items sharing some common characteristics such as ``clothing" or ``phones", which can be used as priors when there is no or little data for the item.

Each user $u$ is interacting with the platform until a clock time point $T$.
This means that each user has a variable number $t_u = t_u(T)$ of interactions with the platform, which is a function of time $T$.
For each interaction $t \le t_u$ the platform orders all possible items in $I$ for the user in a slate $a_t^u$
\begin{equation}
    a_t^u := [\alpha_t^u(1), \alpha_t^u(2), ..., \alpha_t^u(|I|)] \in \perm(|I|),
\end{equation}
where $ \perm(|I|)$ is the set of all permutations of $|I|$ items.
The platform presents the ordered slate $a_t^u$ to the user as a scroll-able list so that $\alpha_t^u(j)$ is the $j^{th}$ item in the list presented to user $u$ at interaction $t$.

\begin{figure}[ht] 
\includegraphics[width=1.0\textwidth]{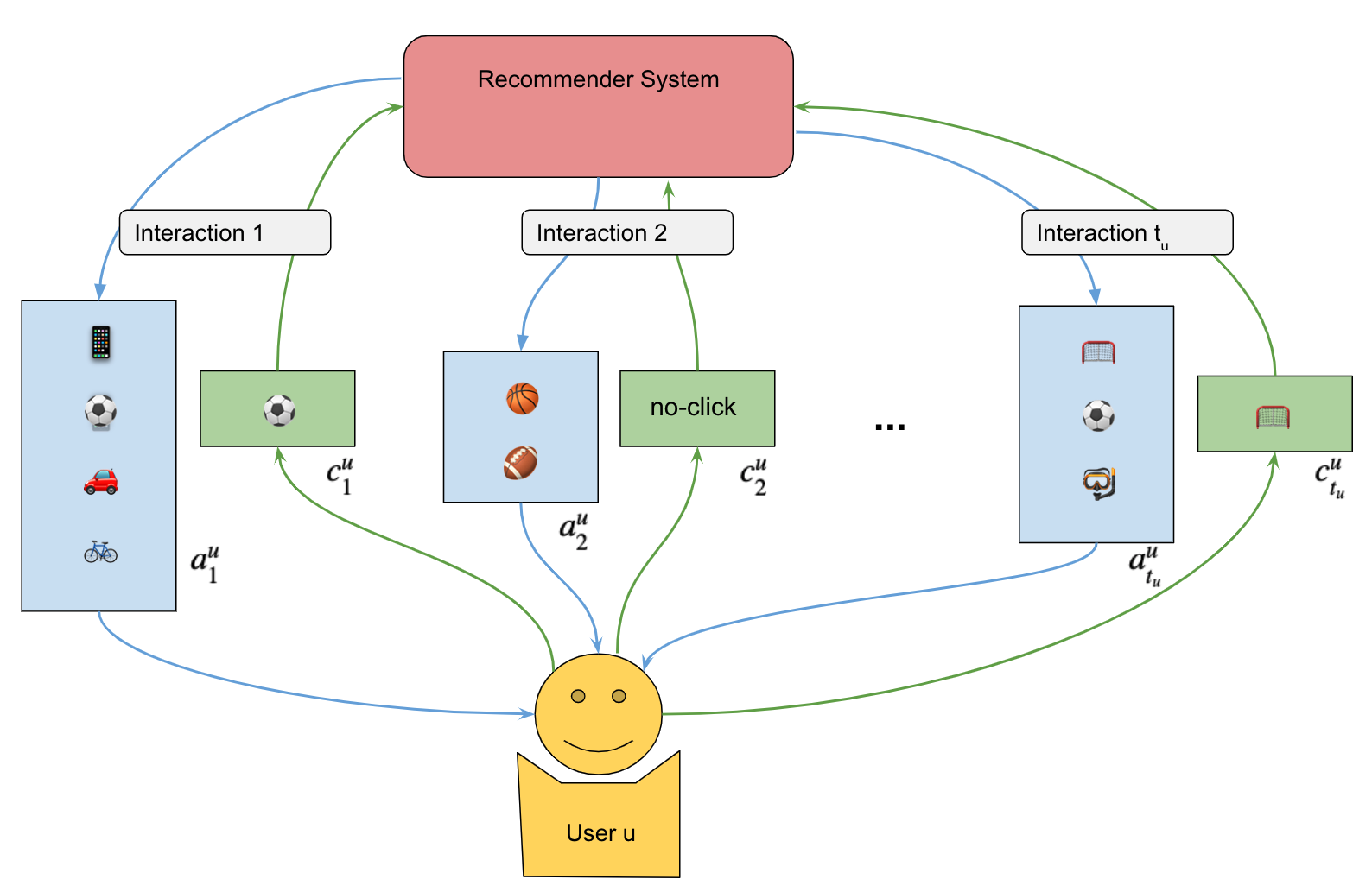}
\caption{Illustration of the interactions between a user and the recommender system: The recommender system is iteratively utilizing its model to recommend slates (blue) to the user. The user, on the other hand, provides feedback in forms of clicks (green).}
\label{fig:interaction_illustration}
\end{figure}

The user may scroll as far as she likes.
Let $s_t^u$ be the total number of items seen by user $u$ at interaction $t$. 
Then the user has seen the set of top $s_t^u$ elements $a_t^u(s_t^u) := [\alpha_t^u(1), \alpha_t^u(2), ..., \alpha_t^u(s_t^u)]$.
Given the set of seen items, the user may click on one of the items in $a_t^u(s_t^u)$ or may not click on any of the items, which we represent as clicking on $i=0$.
Let $c_t^u \in \{a_t^u(s_t^u) \cup 0 \}$ be the item that the user $u$ clicks at interaction $t$.
An illustration of this process is shown in Figure \ref{fig:interaction_illustration}.

The short term objective of the recommender system at time $T$ is to find an ordered list of recommendations $a_{t_u+1}^u$ for each user that maximizes the click probability of the user given her previous interactions.
This can be seen as maximizing an expected reward where the reward is 1 for clicks and 0 for no clicks, namely
\begin{equation} \label{eq:objective}
    max_{a_{t_u+1}^u \in \perm(I)} P(c_{t_u+1}^u \ne 0 | a_{t_u+1}^u, \omega_{T}),
\end{equation}
where $\omega_{T} := \{\omega_{t_u(T)}^u \}_{u=1}^{|U|}$, with $\omega_{t}^u := \{ a_{1:{t}}^u, s_{1:t}^u, c_{1:{t}}^u \}$, is the collection of the information the recommender system has about each user up to time $T$.

The long term objective of the recommender system is to maximize the click probabilities over all future interactions with the users.
A formal expansion of the optimization problem in (\ref{eq:objective}) to multiple horizons would lead to a combinatorial explosion of possible actions.
Instead, we take an heuristic approach to the problem and add exploration to trade off near time reward against reward in future interactions.

\section{The Sequential Slate Model} \label{sec:model}
In this section we build a generative model of the data that we call the Sequential Slate Model and describe Bayesian inference of its parameters.
We will then use this model with its inferred parameters in Section \ref{sec:strategy} to construct recommendations.

There are three types of random variables in the sequential slate problem: 
For each user $u \in U$ at each interaction step $t \le t_u$ we have the ordered slate $a_t^u$ proposed by the platform, the scrolling length $s_t^u$ decided by the user and the item clicked $c_t^u$ by the user.
We condition these random variables on past events to form the slate likelihood
\begin{equation} \label{eq:likelihood_general}
    \begin{split}
    P_{slate}(\omega_{T} | \theta)   =&
    \prod_{u=1}^{|U|} \prod_{t=1}^{t_u} 
    \PT(c_t^u  | \omega_{t-1}^u, a_{t}^u, s_{t}^u) 
    \PT(s_t^u | \omega_{t-1}^u, a_{t}^u) 
    \PT(a_t^u    | \omega_{t-1}^u) \\
\end{split}
\end{equation}
where $\theta$ is a vector of model parameters and $\omega_0 = \emptyset$ is the empty set.
%
We are interested in the user's preferences, and not in the platform's dynamics that generated the slates.
Therefore we assume all the slate random variables as given, i.e. $\PT(a_t^u    | \omega_{t-1}^u) = 1$.
We further assume that the scrolling length $s_t^u$ is a user and history independent Poisson random variable controlled by a parameter 
$\lambda_{s} \in \real$ so that 
$\PT(s_t^u | \omega_{t-1}^u, a_{t}^u) = Poisson(s_t^u; \lambda_{s})$.
This assumption is greatly simplifying, as decisions taken by the users are more complex: the user may either continue to scroll if she has not found what she is interested in yet, or lose interest and stop scrolling.
However, the assumption allows the model to have a variable scrolling length while keeping inference simple. 
In fact, with this assumption we show in Section \ref{sec:strategy} that $\lambda_{s}$ does not need to be estimated as it will not influence the action of the recommender system.

We now specify the first factor in (\ref{eq:likelihood_general}), namely the probability of the user's click given the past behaviour.
There are two sources of information, in addition to scrolling length, that can be used to model the user's behaviour: the previous clicks and the previously shown slates.
We assume that, given the parameters $\theta$, $c_t^u$ depends on the previous clicks made by the user but is conditionally independent of $a_s^u$ and $s_s^u$ for $s < t$:
\begin{equation*}
    \PT(c_t^u  | \omega_{t-1}^u, a_{t}^u, s_{t}^u) =\PT(c_t^u  | c_{1:(t-1)}^u, a_{t}^u, s_{t}^u).
\end{equation*}
We lose information by not modeling the dependence on earlier slates, but we believe that most of the information about the user's preferences lies in the previous clicks made by the user.
Similar model choices were also developed in \cite{Ie2019a}.

To model the user preferences and item characteristics we assume that each can be described by a latent vector in $d$-dimensional space $\real^d$.
Each dimension of the space can be interpreted as a continuous topic, and each region of the space can therefore be interpreted as a set of topic values that describe either a user preference or item.
We assume that each item $i \in I$ has a fixed latent location $v_i \in \real^d$ in this space.
On the other hand, users have dynamic user profiles $z_t^u \in \real^d$ that depend on clicks and  moves in the latent space during interactions with the system.
We model the user profiles as a Gated Recurrent Neural Network \citep[see][]{Cho2014} operating on the click sequence and with user-specific initial locations, which we describe in more detail in Section \ref{sec:userprofile}.
Both the item vector locations $v_i$ and the dynamics of the user profiles $z^u_t$ are learned from data.

For a given user $u$ and interaction $t$, we assume the user 
makes a choice for an item $c_t^u \in I$ from the viewed slate $a_t^u(s_t^u)$ according to
\begin{equation} \label{eq:click-softmax}
    \PT(c_t^u | s_t^u, a_t^u, c_{1:t-1}^u) = 
    \frac{\rel_{t,u, c_t^u}}
        { \sum_{i \in \{a_t^u(s_t^u) \cup 0\}}  \rel_{t,u,i}},
\end{equation}
where 
$\rel_{t,u,i} \in \real^+$ is a relevance function between the user state $z_t^u$ for user u at interaction t and item i.
For all items $i \ne 0$ we let the relevance score be defined by the exponential of the negative Euclidean distance, i.e. $\rel_{t,u,i} = \exp\{-||z_t^u - v_i||\}$.
For the no click item $i=0$ we set the relevance function be equal to a user-independent bias parameter $\rel_{t,u,0} = \beta_{s_t^u}$. 
We allow the bias term $\beta_{s_t^u}$ to depend on the scroll length $s_t^u$.
As the denominator in (\ref{eq:click-softmax}) grows with larger scroll lengths $s_t^u$ we expect $\beta_{s_t^u}$ to be an increasing function in $s_t^u$ to balance the increasing magnitude.

Combining all assumptions above, we write the slate likelihood in (\ref{eq:likelihood_general}) as
\begin{equation} \label{eq:likelihood}
    \begin{split}
    P_{slate}(\omega_{T} | \theta)   =&
    \prod_{u=1}^{|U|} \prod_{t=1}^{t_u} 
    \frac{\rel_{t,u, c_t^u}}
        { \sum_{i \in \{a_t^u(s_t^u) \cup 0\}}  \rel_{t,u,i}}
    Poisson(s_{t}^u; \lambda_{s}).
     \\
    \end{split}
\end{equation}

\subsection{Slate likelihood vs. all-item likelihood} \label{sec:slatevsall-item}
The slate model in (\ref{eq:click-softmax}) assumes that the user only consider what is presented in the slate.
In contrast, the common assumption in recommender systems is that the user considers all items $I$ at each step  (see e.g. \cite{hidasi2016b},
\cite{Hu2008}). 
These models work with an `all-item likelihood' function that sums the relevance scores of \emph{all} items in the denominator:
\begin{equation} \label{eq:likelihood_all-item}
    \begin{split}
    P_{all}(\omega_{T} | \theta)   =&
    \prod_{u=1}^{|U|} \prod_{t=1}^{t_u}
    \frac{\rel_{t,u, c_t^u}}
        { \sum_{i \in I}  \rel_{t,u,i}}
    \end{split}
\end{equation}
The item set $I$ often consists of very many, perhaps millions, of items making this a completely unrealistic assumption, and making the repeated evaluation of the likelihood numerically unfeasible. The common practice is to approximate the denominator by summing over a subsample of $I$, chosen at random. This alleviates the computational problem, but not the implausible modelling assumption that the user considers all items at each interaction. However, the all-item likelihood is often the only option as most available datasets does not contain information about offered slates and scroll lengths $s^u_t$. 
The data presented in this article in Section \ref{sec:dataset} includes full information about which items a user has been presented with.

To compare these modelling approaches, we rewrite the all-item likelihood (\ref{eq:likelihood_all-item}) in terms of the slate likelihood (\ref{eq:likelihood}) and find that the difference is a ratio between the sum of scores of exposures to the sum of scores of all items
\begin{equation} \label{eq:all-item-likelihood}
    \begin{split}
    P_{all}(\omega_{T} | \theta) &= 
    \prod_{u=1}^{|U|} \prod_{t=1}^{t_u}
    \frac{\rel_{t,u,c_t^u}}
        { \sum_{i \in \{a_t^u(s_t^u) \cup 0\}} \rel_{t,u,i} }
    \frac{{ \sum_{i \in \{a_t^u(s_t^u) \cup 0\}} \rel_{t,u,i} }}
        {{ \sum_{i \in I} \rel_{t,u,i} }} \\
    &\propto
    P_{slate}(\omega_{T} | \theta)
    \prod_{u=1}^{|U|} \prod_{t=1}^{t_u}
    \frac{{ \sum_{i \in \{a_t^u(s_t^u) \cup 0\}} \rel_{t,u,i} }}
        {{ \sum_{i \in I} \rel_{t,u,i} }}
    \end{split}
\end{equation}
where we omit the conditionally independent scrolling length $s_t^u$.
For a trained model (i.e.\ with large $P_{all}(\omega_{T} | \theta)$ in (\ref{eq:all-item-likelihood})) we see that the relevance score $\rel_{t,u,i}$ of the all-item likelihood will be high when 
(i) many user's have clicked on an item after being exposed to it,
and (ii) the item has been exposed to many users.
The first condition is the desired objective, whereas the second condition does not reflect any user preference and may not necessarily pull in a constructive direction.
A re-interpretation of (\ref{eq:all-item-likelihood}) considers the second fraction as a generative model of the slates by setting 
$P(a_t^u | \omega_{t-1}) \propto \frac{{ \sum_{i \in \{a_t^u(s_t^u) \cup 0\}} \rel_{t,u,i} }}{{ \sum_{i \in I} \rel_{t,u,i} }}$.
In this interpretation the all-item likelihood assumes that the user's click probability of an item and the platform's probability of recommending the same item are proportional to each other.
When the slate data $a_t^u$ contains information about the user $u$, the presence of this additional likelihood term $\PT(a_t^u |\omega_{t-1}^u)$ can allow the all-item likelihood to make use of more data than the slate-conditioned slate likelihood $P_{slate}(\omega_{T} | \theta)$. 
If the slates in the dataset are informative about the user, for example being generated from search or by a previously trained recommender system, the all-item likelihood can use the information in $a_t^u$ to improve the fit, whereas the slate likelihood cannot do so because it takes the slates as given.
On the other hand, if the slate's data are non-informative on the user, then the all-item likelihood will be misled by the additional likelihood term $a_t^u$. 

We expect an all-item likelihood model to perform strongly in some cases, for example in a platform where most slates originate from searches.
A search engine uses a query from the user to generate a slate, which is a strong signal of the user's interest.
It is therefore natural to believe that slates generated through search are very relevant to describe the user's behaviour.
Consequently, recommendations by a platform that has a large share of its slates generated from (a good) search engine, may not improve when using the slate likelihood instead of the all-item likelihood.

On the other hand, the all-item likelihood can create a feedback loop, or popularity bias \citep{Chaney2017, Abdollahpouri2017}, in the recommender system, promoting already recommended items.
Live recommender systems are usually trained on slates that come from a previous version of the same recommender system.
The second factor in (\ref{eq:all-item-likelihood}) gives larger relevance scores to items that have previously been shown to the user.
This causes the trained model to promote previously recommended items, whether they were clicked or not, leading to a risk of over-recommending poorly performing items. 

\subsection{User profile dynamics} \label{sec:userprofile}
This section describes two user dynamics models for the user profile $z_{t}^u$ at interaction $t$: a linear model and a non-linear Gated Recurrent Unit (GRU) dynamics.
Both can be written in the form 

\begin{equation} \label{eq:userdynamics_general}
    \begin{split}
        h_{t+1}^u= \eta(h_{t}^u, c_t^u; \theta) \\
        z_{t+1}^u = \zeta(h_{t+1}^u; \theta)
    \end{split}
\end{equation}
where $h_t^u$ are hidden user states at interaction $t$ and $h_0^u \in \real^d$ are initial hidden user states that are learned from data.
The two models differ in the choice of the temporal dynamics function $\eta$ and the projection function $\zeta$.
Both functions depend on some learnable parameters, which we will for simplicity define as $\theta_{user} \in \theta$.

\paragraph{Linear user model.}
The linear user dynamics can be written as 
\begin{equation} \label{eq:linearmodel}
\begin{split}
    h_{t+1}^u &= \eta(h_t^u, c_t^u; \gamma) =
    \begin{cases}
        \gamma h_{t}^u + (1-\gamma)  v_{c_t^u} & \text{if } c_t^u \ne 0 \\
        h_{t}^u & \text{if } c_t^u = 0 \\
    \end{cases} \\
    z_{t+1}^u &= \zeta(h_{t+1}^u) = h_{t+1}^u \\
\end{split}
\end{equation}
where $\gamma \in [0,1]$ is a global drift parameter that determines the rate of preference change when the user observes a new item $c_t^u$.
The linear user model assumes that each user drifts in the latent space and updates her preferences towards the item vector of the latest click $v_{c_t^u}$.
If a user is presented with something she does not click on, her preference does not change.

\paragraph{Nonlinear user model (GRU).}
The Gated Recurrent Neural Net user dynamics is defined as
\begin{equation} \label{eq:gru}
    \begin{split}
        z_{t+1}^u &= \zeta(h_{t+1}^u; \theta_{user}) = W_z h_{t+1}^u \\
        h_{t+1}^u &= \eta(h_{t}^u, c_t^u; \theta_{user}) = 
        \begin{cases}
            (\one^{d_h}-\Upsilon_{t+1})^{T} n_{t+1} + \Upsilon_{t+1}^{T} h_{t} & \text{if } c_t^u \ne 0 \\
            h_{t}^u & \text{if } c_t^u = 0 \\
        \end{cases} \\
    \end{split}
\end{equation}
where 
\begin{equation*}
    \begin{split}
        \text{reset gate: } r_{t+1}^u &= \sigma( W_{ir} v_{c_t^u} + W_{hr} h_{t}^u) \\ 
        \text{update gate: } \Upsilon_{t+1}^u &= \sigma( W_{id} v_{c_t^u} + W_{hd} h_{t}^u ) \\ 
        \text{new gate: } n_{t+1}^u &= \text{tanh}( W_{in} v_{c_t^u} + r_{t+1}^{T} (W_{hn} h_{t}) ) \\ 
    \end{split}
\end{equation*}
and where 
$x^T$ is the transpose of a matrix $x$, 
$tanh(x)$ and $\sigma(x) := \frac{1}{1+e^{-x}}$ are the hyperbolic tangent and logistic sigmoid function applied element-wise on a vector $x$ respectively,
$d_h$ is the dimension of the hidden states $h_t^u, r_t^u, \Upsilon_t^u, n_t^u \in \real^{d_h}$
and $\one^{d_h} \in \real^{d_h}$ is a vector of ones.
Note that there are no user-specific parameters, and all learnable weight-parameters are global model parameter vectors used to calculate the transitions of each user with the following dimensions:
$W_{ir}, W_{id}, W_{in}  \in \real^{d_h, d}, 
W_{hr}, W_{hd}, W_{hn}  \in  \real^{d_h,d_h}, 
W_z \in \real^{d, d_h}$.
Gated Recurrent Neural Networks \citep{Cho2014} have shown good performance in sequential recommender systems \cite[e.g.][]{Ludewig2018}.
They allow for more complex dynamics than the linear model (\ref{eq:linearmodel}).

\subsection{Priors} \label{sec:priors}
We need to place priors on the scroll length, and on item and user parameters.
Setting a prior on the scroll length $\lambda_{s}$ is not important as it is irrelevant to decision-making.
Hence we simply set it to an improper uniform over all real values.
We have very little information on the user profile parameters.
The transition parameter matrices in the recurrent neural network are all global, and their priors will likely be washed away quickly because of the large number of users and interactions. 
Therefore we assume that the elements in all matrices are independently normally distributed with mean zero and variance $\sigma_{rnn} \in \real^+$.
The initial hidden user states $h_0^u$ are given an independent normal prior with mean zero and variance $\sigma_{h_0}$.
In the linear model we set the prior for $\gamma$ as a folded Gaussian $P(\gamma) = N(0.5,0.3)$, truncated inside $[0,1]$.
For positive distributions we chose to use folded Gaussians instead of the more natural Gamma distributions due to computational stability and simplicity.
In our attempts, we were not able to estimate stable Gamma parameters as both $(\alpha, \beta)$ went towards infinity.

\paragraph{Hierarchical item prior}
We use the fact that each item $i$ belongs to a known group $g(i)$ to build a hierarchical prior model.
Using a hierarchical item prior reduces the effective parameter space. 
It also allows the model to have some information on new items that have not yet been seen, and thereby address the cold-start problem (\cite{Park2013}, \cite{Kula2015}).
A possible enhancement of this framework, which we do not consider in this article, is to allow the item vectors to depend on other attributes of the items, such as image or text (e.g. \cite{Eide2018}).

For each item $i$ in group $g(i)$ we assume a multivariate normal prior distribution of the parameter vector $v_i$ given its group parameters
\begin{equation}
    P(v_{i} | \mu_{g(i)}^g, \Sigma^g_{g(i)})  \sim N(\mu_{g(i)}^g, \Sigma^g_{g(i)})
\end{equation}
where we parameterize each group $j \in G$  with a mean vector $\mu_j^g$ and a covariance matrix $\Sigma_{j}^g$.

We place a multivariate Gaussian hyper-prior on each group mean vector,  $\mu_j^g \sim N(0_d, \kappa_{\mu} {I_{d^2}})$,
where $\kappa_\mu$ is a scalar hyperparameter, $0_d$ is a vector of zeroes and ${I_d}$ is the identity covariance matrix of dimension $d$.
The group covariance parameter matrix $\Sigma_{j}^g$ is modeled as a diagonal matrix where we place a folded Gaussian prior on each diagonal element $k=1,..,d$ which restricts all values to be positive: $\Sigma_{k,k}^g \sim N_{fold}(0_d, \kappa_{\Sigma} I_{d^2})$, for a hyperparameter $\kappa_{\Sigma}$.
The item prior with its hyperpriors can then be written as
\begin{equation}
\begin{split}
    & P(\{v_i\}_{i \in I},  \{\mu_j^g \}_{j \in G}, \{\Sigma_j^g \}_{j \in G}) \\
    &= P(\{v_i\}_{i \in I} |  \{\mu_j^g \}_{j \in G} , \{\Sigma_j^g \}_{j \in G}) \cdot  P(\{\mu_j^g \}_{j \in G}) \cdot P(\{\Sigma_j^g \}_{j \in G}) \\
    &= \prod_{i \in I} N(v_i; \mu_{g(i)}^g, \sigma_{g(i)}^g)
    \prod_{j \in G} N(\mu_j^g ; 0_d, I_{d^2}) N(\sigma_{j}^g ; 0_d, \kappa_{\Sigma} I_d) .
\end{split}
\end{equation}

We assume that the scroll length, and the user and item prior components are independent of each other, so the full prior distribution is written as
\begin{equation} \label{eq:prior}
\begin{split}
    & P(\theta) = P(\lambda_{s},   \{w \}_{w \in \theta_{user}},   \{v_i\}_{i \in I},    \{\mu_j^g \}_{j \in G}) \\
        &= P(\lambda_{s})  
        \cdot P(\{w \}_{w \in \theta_{user}}) 
        \cdot P(\{v_i\}_{i \in I},  \{\mu_j^g \}_{j \in G}, \{\Sigma_j^g \}_{j \in G}), 
\end{split}
\end{equation}


\subsection{Posterior Approximations} \label{sec:posterior}
The posterior distribution is complex, and we find that a variational approximation is the most efficient way to perform inference.
We use the annealed posterior distribution approach of \citet{Mandt2016,Wenzel2020}, where we temper the likelihood component of the posterior
\begin{equation} \label{eq:posterior}
        P^{(\tau)}(\theta | \omega_{T})  \propto P(\omega_{T} | \theta)^{1/\tau} P(\theta).
\end{equation}
Here 
$\tau \in (0,1]$ is a temperature hyperparameter controlling the relative weight of prior and likelihood in the posterior distribution.
Smaller values of $\tau$ cause the posterior distribution to be more peaked around the maximum likelihood solution.
Tempering the posterior or likelihood has shown to provide better fitted models \citep{Wenzel2020, Mandt2016} than the non-tempered posterior $\tau=1$.

We approximate $P^{(\tau)}(\theta | \omega_{T})$ using Stochastic Variational Inference \citep{Ranganath2014,Blundell2015,Bingham2018}.
Specifically we minimize the Kullback–Leibler (KL)-divergence between the posterior distribution  $P^{(\tau)}(\theta | \omega_{T})$ and an approximate posterior $q_{\phi}(\theta)$ parameterized by a set of parameters $\phi$, which will depend on $\omega_{T}$. 
The optimal parameter $\phi$ which minimises the divergence, is denoted as $\phi^*$.
We use the classical mean field approximation, in which the class of functions $q_{\phi}(\theta)$ over which we optimize consists of distributions in which all parameters are independent Gaussians (except for the initial user states).
That is, for each parameter $\theta_k \in \theta$, we assume the marginal variational distribution to be $N(\mu_{\theta_k}, \sigma_{\theta_k})$, for two appropriate variational parameters $\mu_{\theta_k}$ and $\sigma_{\theta_k}$. 

The parameter $\theta$ includes the initial user state $h_0^u$ for every user $u$ in the system. We find that these initial states are usually strongly determined by the user's first clicks, and hence we can perform inference more successfully by amortizing the variational distribution \citep[as in][]{zhang2018} for $h_0^u$ to be element-wise Gaussian with two amortized functions $\mu_0 = \mu_0(c_{1:t_u}^u, \phi)$ and $\sigma_0 = \sigma_0(c_{1:t_u}^u, \phi)$ to model the mean and variance, respectively.
We let $\mu_0$ be an exponentially decaying weighted average over the item vectors of the user's clicks (excluding the no-clicks) and let the initial variance be approximated by the spread of interest the user has shown:
\begin{equation*}
    \begin{split}
        q_{\phi}(h_0^u|c_{1:t_u}^u, \phi) &= N\left(\mu_0(c_{1:t_u}^u, \phi), \sigma_0(c_{1:t_u}^u, \phi)\right) \\
        \mu_0(c_{1:t_u}^u, \phi) &= 
        \frac{\sum_{t=1}^{t_u} \one_{c_t^u \ne0} w^{1/t} \phi_{V_{c_t^u}}}{\sum_{t=1}^{t_u} \one_{c_t^u \ne0} w^{1/t}} \\
        \sigma_0^2(c_{1:t_u}^u, \phi) &= 
        \frac{\sum_{t=1}^{t_u} \one_{c_t^u \ne0} \{ \phi_{v_{c_t^u}} - \text{average}({\phi_{v_{c_{1:t_u}^u}}})\}^2}{\sum_{t=1}^{t_u} \one_{c_t^u \ne0}} \\
    \end{split}
\end{equation*}
There are two key benefits of amortizing the initial user state. 
Firstly, we reduce the number of variational parameters by $|U| * d^h$, which allows us to scale the algorithm to a larger set of users. 
Secondly, it provides (as we show in Section \ref{sec:strategy}) a fast way to produce an initial state of new users, who have not been seen in the training phase, allowing us to recommend to them without implementing an additional training procedure during the recommendation phase.

Finally we define a hyperparameter $\sigma_{max} \in \real^{+}$ which we use to control the range of the variational distribution.
During optimization, we restrict the standard deviation of each marginal Gaussian distributions to be smaller than $\sigma_{max}$ to avoid an observed exploding variance effect during training.
That is, we require all elements of $\sigma_0^2$ and $\sigma_{\theta_k}$ for all $\theta_k \in \theta$ to be smaller than $\sigma_{max}$.
The final posterior approximation can be written as
\begin{equation} \label{eq:meanfield_posterior}
    q_{\phi}(\theta) = 
    \prod_{\theta_k \in \theta, \theta_k \neq  h_0^{1:U}} N(\theta_k; \mu_{\theta_k}, \sigma_{\theta_k}) 
    \prod_{u=1}^{|U|} N(h_0^u ;\mu_0(c_{1:t_u}^u, \phi), \sigma_0(c_{1:t_u}^u, \phi)),
\end{equation}
where $\mu_{\theta_k}$ and  $\sigma_{\theta_k} < \sigma_{max}$ are the variational parameters which we denote as $\phi = \{\mu_{\theta_k}, \sigma_{\theta_k} \}_{k=0}^{|\theta|}$.
We provide more details on the optimization in appendix \ref{sec:optimization}.

\paragraph{Two-step optimization}
Optimizing the KL divergence for model (\ref{eq:posterior}) for both the item and user parameters simultaneously when using the GRU user model (\ref{eq:gru}) converged to a poorly-performing local optimum. We therefore suggest a two step procedure. The idea is to first infer the item parameters using a relatively simple user dynamic and then inferring the user dynamics in step two while fixing the variational parameters corresponding to the item parameters $v_{1:I}$.
In the first step, we estimate the variational item parameters while using the linear user dynamics  (\ref{eq:linearmodel}), and let the optimization run until the log likelihood of the validation dataset has stopped improving.
The second step continues the optimization from the converged values of step 1 with the following modifications: it uses the GRU user model (\ref{eq:gru}) to produce the user states $z_t^u$ and optimizes over all parameters except the item parameters $v_{1:I}$ which we freeze at the converged values in step 1. 
Again, we let the algorithm run until convergence of the log likelihood on the validation dataset.


\section{Recommendation strategy} \label{sec:strategy}
With a model and an algorithm to perform posterior inference on all unknown parameters, we are ready to move our attention to the recommender system and to determine what recommendation strategy to use.
Combining the objective in  (\ref{eq:objective}) with the approximate posterior distribution $q_{\phi^*}(\theta)$ in  (\ref{eq:meanfield_posterior}), we can write the click probability given a slate $a_{t_u+1}^u$ as an integral with respect to the posterior distribution:
\begin{equation} \label{eq:bandit_objective}
\begin{split}
    &P(c_{t_u+1}^u \ne 0 | \omega_{T}, a_{t_u+1}^u)
    = 1 - P(c_{t_u+1}^u = 0 | \omega_{T}, a_{t_u+1}^u) \\
    =& 1 - \int \sum_{j=0}^{|I|} P(s_{t_u+1}^u = j | \omega_{T}, \theta) P(c_{t_u+1}^u = 0 | \omega_{T},  a_{t_u+1}^u, s_{t_u+1}^u, \theta) P(\theta | \omega_{T}) d\theta \\
    \approx & 1 - \int \sum_{j=0}^{|I|} Poisson(j; \lambda_{s}) \frac{\rel_{t_u+1, u, 0}}{\sum_{i \in 0 \cup a_{t_u+1}^u (j)} \rel_{t_u+1, u, i}}   q_{\phi^*}(\theta) d\theta
\end{split}
\end{equation}
where in the the last step we used (\ref{eq:likelihood}).
Equation (\ref{eq:bandit_objective}) is intractable due to the multidimensional integral over all parameters $\theta$.

Conditional on $\theta$, the optimization of (\ref{eq:bandit_objective}) can be reduced to a ranking problem of items $i$ based on the relevance scores $\rel_{t_u+1, u, i}$.
This can be seen in two steps. Firstly, given a scroll length $j$, a slate $a^u_{t_u+1}$ which optimizes (\ref{eq:bandit_objective}) places the $j$ items with the highest relevance scores in the first $j$ places, where the order between them does not matter.
Secondly, since this is true for all $j \le |I|$, an $a^u_{t_u+1}$ that orders all items from highest to lowest relevance score will maximise (\ref{eq:bandit_objective}) for all realizations of $s_{t_u+1}^u$.

Several recommendation strategies are possible, and here we present three: Greedy, Single Thompson sampling and a variant of Thompson sampling we call in-slate Thompson sampling.
Common to all three strategies is that they are efficient to compute, in contrast to the Upper Confidence Bound (see e.g. \cite{Lattimore2019}, ch 7) which would have required multiple evaluations of the integral in (\ref{eq:bandit_objective}) and therefore would be computationally unfeasible in a recommender setting.

\begin{strategy}[Greedy maximum a posteriori (greedy MAP)] \label{strategy:greedy}
\leavevmode
\begin{enumerate}
\item Take the maximum a posterior probability estimate:  $\theta_{MAP} = \text{argmax}_{\theta} q_{\phi^*}(\theta) $. 
\item Construct $a_{t_u+1} ^u$ by ranking all items $i$ according to their relevance score $r_{\theta_{MAP}}(t_u+1, u, i)$.
\end{enumerate}
\end{strategy}
Step 1 is easy to compute in our mean-field (approximate) posterior, as the model parameters are marginal Gaussian in the variational approximation $q$ and the maximum a posterior is the mean of the variational distribution.
Recommendation strategy \ref{strategy:greedy} does not recognize that $\theta$ is an estimate with uncertainty and is using the a posteriori maximum estimate as a point estimate.

In sequential recommender systems, although there are large amounts of data in general, there is often little data per user and per item.
This makes the posterior distribution of the user initial condition $h_0^u$ and item vectors $v_{1:I}$ uncertain and rather flat.
Thompson sampling is an algorithm that samples one set of parameters from the posterior in (\ref{eq:posterior}), and then acts greedily.
It has been found to be a simple and efficient way of balancing the exploitation of immediate performance and exploration of the parameter space for future exploitation \citep{Russo2017}.
The randomization in the posterior sampling ensures exploration when and where it is worth exploring. 
Quoting \citet{Lattimore2019}, 
\textit{
"If the posterior is poorly concentrated, then the fluctuations in the samples are expected to be large and the policy will likely explore. On the other hand, as more data is collected the posterior concentrates towards the true environment and the rate of exploration decreases."
} 
In our scenario this may be implemented in the following way:
\sloppy
\begin{strategy}[Single Thompson Sampling] \label{strategy:singleTS}
For each user state 
$\omega_t^u = \{a_{1:t}^u , c_{1:t}^u\}$ do the following:
\begin{enumerate}
    \item Sample one set of parameters $\tilde{\theta} \sim q_{\phi^*}(\theta)$ from  (\ref{eq:posterior}).
    \item Construct $a_{t_u+1} ^u$ by sorting all items $i$ according to the relevance score $r_{\tilde{\theta}}(t_u+1, u, i)$.\end{enumerate}
\end{strategy}
Note that the additional effort over greedy MAP is simply a single sample from $q_{\phi^*}(\theta)$.

In the third recommender strategy, that we call in-slate Thompson Sampling, we sample our posterior for each position within each slate.
The degree of exploration is larger than for Single Thompson Sampling, and we introduce diversity into every slate instead of diversifying between slates.
In this way we can increase the probability that the user finds something interesting, thus increasing the click rate on every slate.
This idea is similar to some other recent suggestions, e.g. \cite{Edwards2018}.
\begin{strategy}[In-slate Thompson Sampling] \label{strategy:inslateTS}
Define a number of sampling parameters $J$. 

\begin{enumerate}
    \item Sample $J$ sets of parameters $\tilde{\theta_j} \sim q_{\phi^*}(\theta)$ from (\ref{eq:posterior}).
    \item For each sample $j=1,2,..,J$, compute $a_{j, t_u+1} ^u$ by sorting all items $i$ according to the relevance score $r_{\tilde{\theta}}(t_u+1, u, i)$.
    \item Construct a final recommendation list $a_{j, t_u+1} ^u$ by displaying the first recommendation from each sample, then the second from each sample and so on. If item i has already been added to the list, it is omitted.
\end{enumerate}
\end{strategy}
This third strategy is more involved than the previous two, but by carrying out slate optimization for several different sampled $\theta$ values we expect to see more diverse slates when the posterior distributions are not concentrated.

\section{A new marketplace dataset with views and clicks} \label{sec:dataset} \label{sec:dataset_finn}
We have collected a dataset as described in Section \ref{sec:problem} from the Norwegian online marketplace FINN.no. 
This article makes an anonymized version of these data publicly available for research purposes, which is available at \href{https://github.com/finn-no/recsys-slates-dataset}{https://github.com/finn-no/recsys-slates-dataset}.
FINN.no is the leading marketplace in the Norwegian classifieds market and provides users with a platform to buy and sell general merchandise, cars, real estate, as well as house rentals and job offerings.
The dataset consists of users' behaviour for 30 days, logging the slate interactions each user had with the site.
For each user $u$ and interaction step $t$ we recorded all items in the visible slate $a_t^u(s_t^u)$ (up to the scroll length $s_t^u$), and the user's click response $c_t^u$.
We also logged whether the slate presented originated from a search query or recommendation.
Approximately 80\% of the slates shown come from a search query, and 20\% are suggested by the recommender systems.
The dataset consists of $37.4$ million interactions,  $|U| \approx 2.3$ million  users and $|I| \approx 1.3$ million items that belong to one of $|G| = 290$ item groups.

\paragraph{Anonymization and reductions}
The dataset has been reduced to better anonymize users and to make it easier to handle computationally. 
First, since users are presented with very many slates that are never clicked, we uniformly sample and keep only 10\% of these non-clicked slates in the dataset.
The second reduction is to ensure some balance across users: we record the first $t_u=20$ interactions for each user in the dataset and do not consider users with less than 10 interactions.
Finally, the total number of items a user may view at any given interaction is capped by $25$, which only affects approximately 5\% of the slates; 
any item appearing further down the slate is omitted, and any interaction that led to clicks on these items is removed.

\paragraph{Training, validation and test datasets}
We split the data into training, validation and test datasets in the following way: 
First let $t_u^{test}$ be the maximum number of interactions a user has recorded in the dataset.
Second, we uniformly sample 90\% of the users, and assume that all their recorded interactions occurred before the (wall) time threshold $T$: $t_u(T) = t_u^{test}$.
All these interactions are placed in the training dataset.
For the remaining 10\% of the users we assume that the first 5 interactions happened before time $T$ and include these interactions in the training dataset $t_u = 5$ as well.
By keeping the first five interactions from each user we are able to estimate the initial user state to some degree, thus simplifying the cold start problem.
All interactions above $t_u(T)$ are placed in validation and test sets with 5\% of the total users in each.
Each user will then have $t_u^{test} - t_u$ interactions in the test and validation datasets, respectively. 
We denote the sets of users in the three datasets as $U_{\text{train}}$, $U_{\text{test}}$ and $U_{\text{valid}}$.

\section{Experiments} \label{sec:experiments}

Section \ref{sec:model} motivates the investigation of multiple hypotheses:
\begin{enumerate}
    \item Does a recommender system improve when using the more realistic slate likelihood (\ref{eq:likelihood})
    instead of the all-item likelihood (\ref{eq:all-item-likelihood})? 
    \item Is it beneficial to use hierarchical priors (Section \ref{sec:priors}) when modeling items?
    \item Is the non-linear user dynamics such as a Gated Recurrent Neural net needed to accurately model the user, or is a linear dynamics sufficient? 
    \item Do recommender strategies that exploit uncertainty improve recommendations as a result of increased diversity and improved learning, or do they suffer because of the non-optimizing actions taken during exploration?
\end{enumerate}
We investigate the first three hypotheses in three ways: 
(1) an offline experiment (Section \ref{sec:offlineexperiment}) using the dataset described in Section \ref{sec:dataset_finn},
(2) an online experiment (Section \ref{sec:onlineexperiment_model}) with real users in an internet marketplace,
and (3) a natural online experiment (Section \ref{sec:natural_experiment}) that shows how the performance of the slate likelihood model and the all-item likelihood model change when training data do not include a large proportion of slates $a_t^u$.
The fourth hypothesis is evaluated in an online experiment (Section \ref{sec:onlineexperiment_strategy}), as measuring exploration-effects is extremely challenging offline, especially in a slate-based recommender setting, since in an offline experiment we can not observe how a user would have responded to a different or more diverse slate.
Finally we summarise the experimental results in Section \ref{sec:analysis}.

\subsection{Offline model experiment} \label{sec:offlineexperiment}
The offline experiment evaluates the slate likelihood model with GRU and item hierarchy against three other variations where we have changed one important feature of the model: 
an all-item likelihood variant, a linear user profile variant and a variant without item hierarchical priors.
A detailed description of these variants can be found in Table \ref{tbl:offline-results}.

We report two evaluation metrics: the log likelihood of the model corresponding to the test dataset and Hitrate@K for $K=20$.
In this article we define the commonly used hitrate metric in the following way:
For each user $u$ in the test set $U_\text{test}$ who has interacted with the platform up to interaction $t_u$,
we let the recommender system generate a set of $K>0$ recommendations $a_{t_u+1}^u(K)$ using the greedy MAP approach (Recommender Strategy \ref{strategy:greedy}).
The Hitrate@K is defined as the average size of the intersection between the recommended items $a_{t_u+1}^u(K)$ and the observed clicks of the user in the test data:
\begin{equation} \label{eq:hitrate}
    \text{Hirate@K} = \frac{1}{|U_{\text{test}}|} \sum_{u \in U_{\text{test}}} | a_{t_u+1}^u(K) \cap c_{(t_u+1):{t_u^{test}}}^u|
\end{equation}

We run an extensive hyperparameter search for all models over the most relevant hyperparameters using Bayesian Optimization on the validation hitrate.
For example, we vary the width of the variational distribution $\sigma_{max}$ and the relative weight $T$ of the likelihood versus the prior.
When $T$ and $\sigma_{max}$ goes to zero, then the solution corresponds to the frequentist maximum likelihood solution, which is therefore considered as an option in the Bayesian Optimization.
The methodology is described in detail in Appendix \ref{sec:hypertune}.

The results of the offline experiment are shown in Table \ref{tbl:offline-results}.
Note that the likelihood of the all-item-likelihood model is different from and therefore not comparable with the likelihood of the other models. 
The all-item likelihood model is, however, comparable with the others in terms of hitrate.
In terms of test log likelihood we see that models using a gated recurrent neural net for user dynamics and hierarchical item priors outperforms variants that do not use these features.
However, in terms of hitrate, we see that the linear user dynamics model performs better than all others.
The all-item likelihood model outperforms the slate likelihood model in terms of hitrate.

\subsection{Online model experiment} \label{sec:onlineexperiment_model}
\begin{figure} 
    \includegraphics[width=\textwidth]{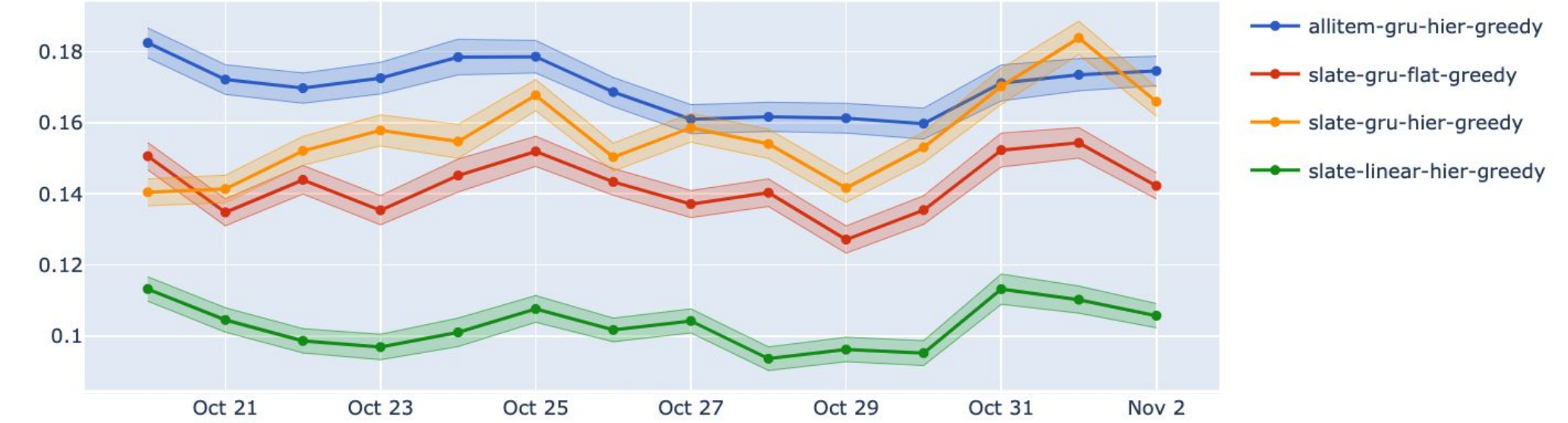}
    \caption{Click rates per model variant and day in the online experiment.}
\label{fig:online-greedy-daily}
\end{figure}

We evaluate the same four models considered in Section \ref{sec:offlineexperiment}, but on real users of an internet marketplace platform.
We again use the greedy MAP approach to select actions (Recommender Strategy \ref{strategy:greedy}) since the aim is still to test whether the model predicts user actions better.
In the experiment, each algorithm was given an equal (random) share of the user traffic of the online marketplace over a period of 14 days. 
Each user was allocated to the same model for the entirety of the test.
We use click rates of each model variant as the evaluation metric of the test.
The click rate for a specific model over a period of time is the total number of clicks made divided by the total number of slates shown.
The results are reported in Table \ref{tbl:online-greedy}, and are also visualized over time in Figure \ref{fig:online-greedy-daily}.
We see that in terms of click rates on real users the all-item likelihood model outperforms the other variants, and using hierarchical priors and a gated recurrent neural net in the user dynamics outperforms the non-hierarchical and linear variants, respectively.

\begin{table}
\caption{
Result of the offline and online experiments.
The four models are named using four keywords: 
(1) \emph{Slate} or \emph{all-item} denotes whether the slate likelihood in  (\ref{eq:likelihood_general}) or the all-Item likelihood in (\ref{eq:all-item-likelihood}) is used; 
(2) \emph{gru} or \emph{linear} denotes whether we use the GRU user dynamics described in  (\ref{eq:gru}) or the linear dynamics in (\ref{eq:linearmodel}); 
(3) \emph{hier} or \emph{flat} denotes whether we use the hierarchical item priors in Section \ref{sec:priors} or a non-informative prior $P(v_i) = N(0, \sigma_v)$ with hyperparameter $\sigma_v=0.1$; 
(4) \emph{greedy} implies that we use the maximum a posteriori Recommendation Strategy \ref{strategy:greedy}.
Note that \emph{all-item-gru-hier-greedy} has a different likelihood function and can not be compared with the others.
We have used $K=20$ when calculating hitrate from (\ref{eq:hitrate}).
In the offline experiment, in terms of hitrate, we see that the linear variant performs best. In the online experiment, in terms of click rate, the all-item likelihood variant is best.}

\begin{tabular}{c | c c c c c}
\hline\noalign{\smallskip}
Algorithm & Test log likelihood & Offline hitrate & Online click rate \\ [0.5ex] 
\noalign{\smallskip}\hline\noalign{\smallskip}

Slate-gru-hier-greedy    & -4.96e+07        & 0.105         &  15.6\% \\ 
Slate-linear-hier-greedy & -4.98e+07        & \textbf{0.149}&  10.3\% \\
Slate-gru-flat-greedy    & -4.97e+07        & 0.0988        &  15.5\% \\
all-item-gru-hier-greedy  & -2.81e+07        & 0.125         & \textbf{17.0\%} \\
\noalign{\smallskip}\hline

\end{tabular}
\label{tbl:online-greedy} \label{tbl:offline-results}
\end{table}

\subsection{Natural online comparison between slate likelihood and all-item likelihood} \label{sec:natural_experiment}
During the online experiment an accidental bug in the tracking system was introduced causing no search slates to be recorded correctly for a full week.
The bug created a natural experiment to investigate our hypothesis in Section \ref{sec:slatevsall-item} that the performance of the all-item likelihood model is enhanced by the presence of slates $a_t^u$ that are particularly relevant to the user $u$.
On this platform approximately 80\% of all slates in the training dataset are search slates.
A loss of one week of search data in a training dataset that collects the last four weeks for training implies that we reduce the proportion of search slates to 75\% in the training dataset.
Since we believe the all-item likelihood model to learn from informative slates (Section \ref{sec:slatevsall-item}), we expect slate likelihood to perform better compared to the all-item likelihood in the period when there is less search slates in the training data.

\begin{figure}
\includegraphics[width=\textwidth]{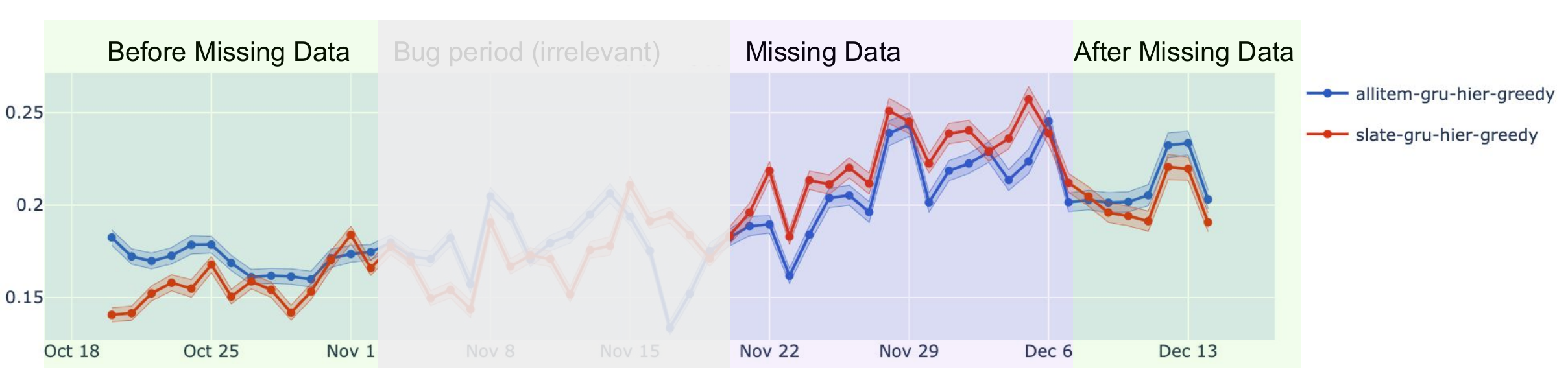}
\caption{Experiment showing the performance of \emph{Slate-gru-hier-greedy} and \emph{all-item-gru-hier-greedy} before, during and after the loss of 25\% of the search data on the marketplace platform. It also displays the period with the corresponding bug (in grey) when measures are faulty and should not be considered, but are given for completeness.
The figure shows that the slate likelihood model performs better than the all-item likelihood model when the dataset consists of slates $a_t^u$ that are less relevant to the user $u$.
}
\label{fig:naturalexperiment}
\end{figure}

In Table \ref{tbl:slatevsall-item} we compare the performance of \emph{Slate-gru-hier-greedy} and \emph{all-item-gru-hier-greedy} before, during and after the period of the missing search data.
We also visualize the performance of both models over time in Figure \ref{fig:naturalexperiment}.
As expected, we see that the all-item likelihood model outperforms the slate likelihood model when we have full data history, while the opposite is true when the search data are missing.
This supports our hypothesis that the all-item likelihood model is implicitly modeling the generated slates in addition to the behaviour of the users.

\begin{table}
\caption{Experiment showing results when informative slates were missing. Click rate figures are not comparable over time due to external changes such as seasonal effects, design changes, etc. Relative click rate is shown for comparison over time periods and is defined as the ratio of the click rates between \emph{all-item-gru-hier-greedy} and \emph{Slate-gru-hier-greedy}}
\label{tbl:slatevsall-item}       
\begin{tabular}{l|rrr}
\hline\noalign{\smallskip}
Model & Before Missing Data & Missing Data &  After Missing Data  \\
\noalign{\smallskip}\hline\noalign{\smallskip}
Rel. click rate [all-item/ Slate]       & 109.2\% & 93.7\% & 105.5\% \\
\hline
Click rate slate likelihood            & 15.6\% &  22.0\% & 20.2\% \\
Click rate all-item likelihood         & 17.0\% &  20.6\% &  21.3\% \\

\noalign{\smallskip}\hline
\end{tabular}
\end{table}

\subsection{Online recommendation strategy experiment} \label{sec:onlineexperiment_strategy}

We conduct an online experiment to evaluate the performance of the three recommender strategies defined in Section \ref{sec:strategy}.
All three recommender strategies were tested for each of the two model variants \emph{slate-gru-hier-greedy} and \emph{all-item-gru-hier-greedy}, resulting in a total of 6 recommender variants.
Each algorithm was given an equal (random) share of the user traffic of the online marketplace over a period of 5 days, and each user was allocated to the same recommender variant for the entirety of the test.
Every night the models were retrained using all the latest data that had been collected during the previous day.
This implies that all variants benefit from the exploration done by the explorative Recommendation Strategies (Strategies \ref{strategy:singleTS} and \ref{strategy:inslateTS}).
It is therefore not possible to estimate any gain in click rates that comes from exploration.
We estimate click rates for each of the variations to evaluate their performance.
The results are shown in Table \ref{tbl:online-strategy}.

We see that exploration does not hurt click rate performance.
Indeed the most explorative approach, in-slate Thompson Sampling, appears to result in a higher click rates (at least when paired with all-item likelihood), presumably due to diversity giving us a better coverage of user preferences \citep[see][]{Edwards2018}.

\begin{table} 
\caption{
Results of the online Recommendation Strategy experiment.
We keep the naming convention as in Table \ref{tbl:offline-results}.
\emph{singleTS} means that the variation uses recommender strategy \ref{strategy:singleTS},
and \emph{inslateTS} means the use of recommender strategy \ref{strategy:inslateTS}.
}

 \begin{tabular}{c | c c c}
\hline\noalign{\smallskip}
 Algorithm & No. slates (1000) & No. clicks (1000) & Click rate \\ [0.5ex]
\noalign{\smallskip}\hline\noalign{\smallskip}
 Slate-gru-hier-greedy              & 66.3 & 13.4  & 20.2\% \\ 
 Slate-gru-hier-singleTS            & 67.4 & 14.1  & 20.9\% \\ 
 Slate-gru-hier-inslateTS           & 67.9 & 13.6  & 20.0\% \\ 
\\
 all-item-gru-hier-greedy    & 98.0 & 20.5 & 21.3\% \\
 all-item-gru-hier-singleTS & 100.7 & 20.6 & 21.0\% \\
 all-item-gru-hier-inslateTS & 99.6 & 21.9 & \textbf{22.7\%} \\
\noalign{\smallskip}\hline
\end{tabular}
\label{tbl:online-strategy}
\end{table}

\subsection{Discussion} \label{sec:analysis}
We discuss the experimental results in the light of each of the four hypotheses listed in the start of the section.

\paragraph{Slate likelihood vs all-item likelihood}
The all-item likelihood model outperforms the slate likelihood model in the offline test, and in both of the controlled online experiments in sections \ref{sec:onlineexperiment_model} and \ref{sec:onlineexperiment_strategy}.
However, during the missing search-slates period of the online experiment, described in Section \ref{sec:natural_experiment}, the slate likelihood model outperforms the all-item likelihood.
This supports the understanding formalised in Section \ref{sec:model} that the all-item likelihood exploits the information about users in the slates $a_t^u$ to improve the estimation of click probabilities.
In our dataset (Section \ref{sec:dataset_finn}) most of the slates originate from search queries so the $a_t^u$ are highly informative about the users' preferences. 
The experiments therefore indicate that the all-item likelihood is beneficial when users are exposed to informative slates, whereas the slate likelihood is superior when the slates shown to the user are less informative.
We observed that the slate likelihood outperformed the all-item likelihood when the ratio of search slates to recommended slates fell from 80\% to 75\%. 

\begin{figure}[ht] 
\includegraphics[width=1.0\textwidth]{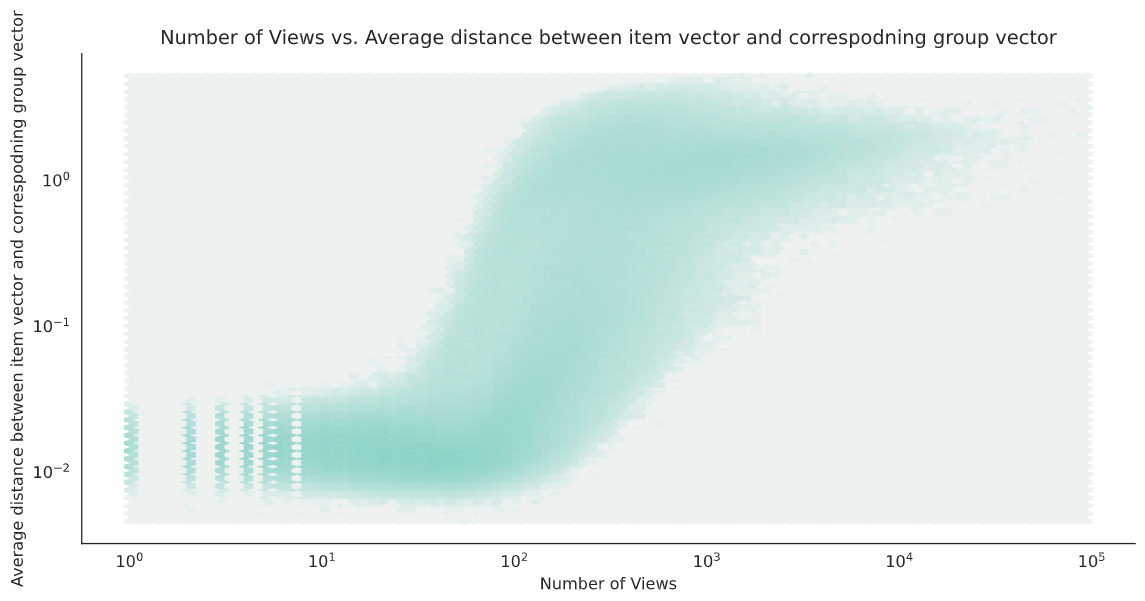}
\caption{Two dimensional histogram of all items in \emph{Slate-gru-hier-greedy}: Number of views vs. difference between item mean vector and the items corresponding group mean vector: $||E_{\phi}(v_i) - E_{\phi}(\mu_{g(i)}^g)||_2$}
\label{fig:prior_vec_dist}
\end{figure}

\begin{figure}[ht]
\includegraphics[width=1.0\textwidth]{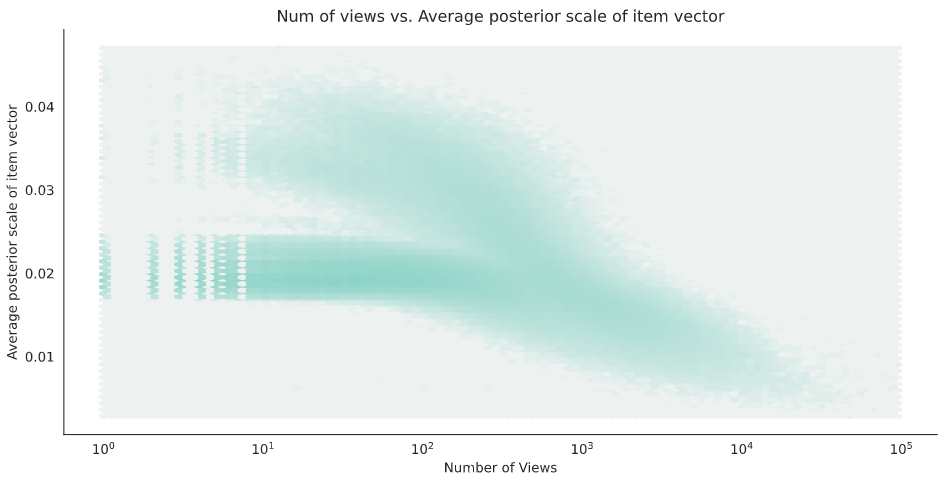} 
\caption{
Two dimensional histogram of all items in \emph{Slate-gru-hier-greedy}: 
The number of views vs. the estimated posterior scale parameter $\sigma_{\theta}$.
We see that items with fewer views have wider posterior distributions.
Darker colour means that more items are present in that histogram basket.} 
\label{fig:item_scale}
\end{figure}

\paragraph{Hierarchical item priors increase click rates}
Both the offline and online experiments in Sections \ref{sec:offlineexperiment} and \ref{sec:onlineexperiment_model} confirm that the use of hierarchical item priors, based on categorical information about the items, improves the recommendation model.
The improvement is small but consistent over all metrics.
Figure \ref{fig:prior_vec_dist} shows the distance between the expected item vector $v_i$ and its corresponding expected group vector for the gru-hier model: $||E_{\phi}(v_i) - E_{\phi}(\mu_{g(i)}^g)||_2$.
As expected, parameters from items with fewer views are closer to their group vectors because we lack more detailed information.
Similarly, Figure \ref{fig:item_scale} shows that there is a negative relationship between the number of views of an item and the posterior estimate of the scale parameter $\sigma_{\theta}$ for the item.

Using hierarchical item priors can therefore be seen as an effective way to incorporate content information when faced with the item cold-start problem. 
Our results also indicate that the item parameters will gradually shift from following prior knowledge to being well estimated as more traffic data on the item is collected.

\paragraph{Linear user dynamics}
The linear user dynamics model \emph{Slate-linear-hier-greedy} is the best performing model according to the hitrate metric, but worst in both log likelihood and click rates as seen in Table \ref{tbl:offline-results}.
The fact that the non-linear slate likelihood models outperform the linear model both in terms of likelihood and click rates indicates that the hitrate metric is not particularly useful for evaluating recommendation performance.
Hitrate@K measures the ability of the recommender system to find the items the user will click in the next $t_u^{test}-t_u$ interactions, based on the user's first $t_u$ clicks but \emph{not considering what slates is presented to the user after $t_u$}.

In contrast to the non-linear GRU, a linear model cannot respond quickly enough to a sudden change in user behavior. 
This makes a linear model less precise than a GRU model but with the potential benefit that it can recommend items that are relevant for additional future interactions.
Conversely, a GRU model may be very precise at each interaction and quickly update its belief about the user preference after each interaction.
The flexibility of the GRU model is therefore not measured by the Hitrate@K metric, and the GRU model may therefore perform badly.
Hence this experiment suggests that hitrate is a misleading metric as it will give a disadvantage to models that quickly updates its user profile. It also shows that it is important to use recommender systems that allow user preference to change during interactions.

\paragraph{Equal or increased performance with explorative recommender strategies}
The online recommender strategy experiment reported in Table \ref{tbl:online-strategy} shows that the recommendation strategies that use Thompson Sampling-style model uncertainty exploration perform equal to or better than their greedy variants.
Decision theory literature argues that there is an explore-exploit trade-off as exploration requires the system to recommend items that are believed to be less relevant \citep{Lattimore2019}. 
In our experiments, since all strategies are given the data from all the other strategies to learn, we are not able to measure any benefit to exploring.
However, the best performing recommendation strategy in our experiments is using the in-slate Thompson sampling.
This shows that adding exploration may increase the click rates of a recommender system even without gaining any learning benefit.
We believe this is due to the added diversity included in each slate, and it confirms a similar claim made in \cite{Edwards2018} that diversity does not need to be a separate objective of a recommender system, but can emerge as a natural consequence of optimizing click rates.

\section{Conclusion} \label{sec:conclusion}

This article considers sequential recommender systems on slates of items and proposes a Bayesian gated recurrent neural network with hierarchical item priors and two explorative recommender strategies.
Accompanying the article, we publish a dataset from a web-based marketplace in Norway that contains not only the clicks but also the slate exposures of the users.

First, we argue that using the common likelihood where the user considers all items in each interaction (all-item likelihood) is an unrealistic model of how users select items to click, and propose an alternative likelihood where the user only considers the slate of items actually presented to her (slate likelihood).
We find that the all-item likelihood can be seen as implicitly modeling the click probability and the slate generation process together. 
We show that despite the unrealistic assumption, the all-item likelihood can fruitfully use information in the slate generation process if the majority of slates presented to the user are informative.
Both our offline and online experiments show that the all-item likelihood performs better than the slate likelihood when the slates presented to the user are informative, and that the opposite is true when the slates are not.
Future work in this area is to model the slate generation process explicitly to provide greater inferential precision than simply combining the slate generation and click process through the all-item likelihood; especially in the case of multiple slate generation processes (search and recommendations), they should be modeled separately.
Second, thanks to the fully probabilistic model and a good approximation of the posterior distribution, we studied two Thompson sampling type explorative recommendation strategies that both performed on par or better than their greedy counterparts.
This shows that there may not be a trade-off between exploration and exploitation in slate (multi-item) recommendation systems, and that adding smart exploration and diversity in slates may also improve click rates.
Lastly, our experiments show that using hierarchical item priors based on item groups or categories can improve click rates, and that using non-linear user dynamics of interactions is superior to using linear dynamics. 

\section*{Declarations}
\paragraph{Funding}
We acknowledge funding from BigInsight (Norwegian Research Council project number 237718) 
and the joint Industrial PhD project between FINN.no, University of Oslo and the Norwegian Research Council (Norwegian Research Council project number 294330).
This research is part of the strategic partnership between BigInsight at the University of Oslo and the STORi Centre for Doctoral Training at Lancaster University.

\paragraph{Code availability}
The code and the data used in the article are available in the following repository: 
\href{https://github.com/finn-no/recsys-slates-dataset}{https://github.com/finn-no/recsys-slates-dataset}.

\paragraph{Conflict of interest}
The authors declare that they have no conflict of interest.

\bibliographystyle{spbasic}
\bibliography{references}

\appendix

\section{Hyperparameter search} \label{sec:hypertune}
The model depends on several hyperparameters, and too many to do an exhaustive search of the hyperparameter space. 
We have identified those that we believe would have greatest impact on the result, and described them with the search ranges in Table \ref{tbl:hyperparameters}.

\begin{table}[ht]
\caption{
List of all recommender systems in the model variation experiment. All recommender systems use Recommender Strategy \ref{strategy:greedy}. 
}
\begin{tabular}{  l |  p{7cm} }
\hline\noalign{\smallskip}
\textbf{Hyperparameter}      & \textbf{Description}   \\
\noalign{\smallskip}\hline\noalign{\smallskip}
    $\sigma_{MAX} \in [0.05,10.0]$      & A cap on the maximum scale of the approximate posterior parameters \\
    $\tau \in (0,1]$                    & Likelihood temperature \\
    $\lr \in [1e-4,1e-1]$               & The learning rate in equation (\ref{eq:gradient_update}) \\
    $\sigma_{rnn}^2 \in [0.1, 10.0]$    & The prior standard deviation of the RNN parameters \\
    $\kappa_{\mu} \in [0.1, 2.0]$       & The prior standard deviation of the group vectors \\
    $\kappa_{\Sigma} \in [0.01, 2.0]$   & The prior standard deviation of the shared scaled per group \\
\noalign{\smallskip} \hline
\end{tabular}
\label{tbl:hyperparameters}
\end{table}
For each algorithm that is tuned, we perform a Bayesian Optimization search using Botorch \cite{Balandat2020} and Ax \footnote{https://github.com/facebook/Ax}.
We use the standard recommendations and settings in Ax, and for each model the Bayesian Optimization algorithm gets a 20 step budget to optimize validation hitrate, sampled on over 100k users.
To obtain a fair amount of resources to all algorithms, each run is constrained by the number of times it can evaluate the data.
We set $E*S \le 800$ which gives a total running time of approximately 10 hrs on the full dataset on one Nvidia P100 GPU.
However, the all-item likelihood is computationally slower and was therefore reduced to 400 to make it comparable with respect to computational budget \footnote{This model was more than twice more time consuming in our experiments, but we only reduced it with 50\% of iterations.}.
The results of the hyperparameter search is given in table \ref{tbl:hyperopt-results}.

\begin{table} 
\caption{Offline test results with optimal hyperparameters. The algorithm names follow the same standard as in Table \ref{tbl:offline-results}, except that we removed the recommender strategy as it is not applicable. Note that the all-item likelihood function is different from the slate likelihood, and this metric is therefore incomparable for the all-item model.}
\begin{tabular}{l|rr|rrrrrr}
\hline\noalign{\smallskip}
        name &  Loglik test &  Hitrate test &  $\sigma_{max}$ &    $T$ &      $l$ &  $\sigma_{rnn}$ &  $\kappa_\mu$ &  $\kappa_{\Sigma}$ \\
\noalign{\smallskip}\hline\noalign{\smallskip}
slate-gru-hier &    -4.96e+07 &         0.105 &            9.22 & 0.0265 & 0.000434 &            7.14 &        4.63 &            0.817 \\
slate-linear-hier &    -4.98e+07 &         0.149 &            7.18 & 0.0339 &  0.00075 &            11.7 &        0.01 &             0.01 \\
slate-gru-flat &    -4.97e+07 &        0.0988 &             5.3 & 0.0096 & 0.000905 &            8.55 &        8.88 &             7.19 \\
all-item-gru-hier &    -2.81e+07 &         0.125 &            2.37 & 0.0736 &  0.00107 &             6.8 &        13.7 &               10 \\
\noalign{\smallskip}\hline
\end{tabular}
\label{tbl:hyperopt-results}
\end{table}

We see that the slate-gru-hier-greedy has achieved the highest test likelihood among the slate likelihood models, and slate-linear-hier-greedy achieves the best hitrate.
Slate-gru-flat-greedy and all-item-gru-hier-greedy has a large $\kappa_{\Sigma}$ compared to the others.
We believe this can be explained by different reasons for the two algorithms: 
The slate-gru-flat-greedy does not have access to group priors, and should therefore have larger distances between items in the same (one) group. 
The all-item-gru-hier-greedy have all items in the denominator of its function, and will therefore record more data on all items than the slate models. 
This cause the posterior distribution to be more peaked for all items.

\begin{figure}
\caption{Bayesian Optimization: Marginal effects of likelihood tempering $\tau$ vs. $\kappa_{\mu}$}
\centering
\includegraphics[width=1.0\textwidth]{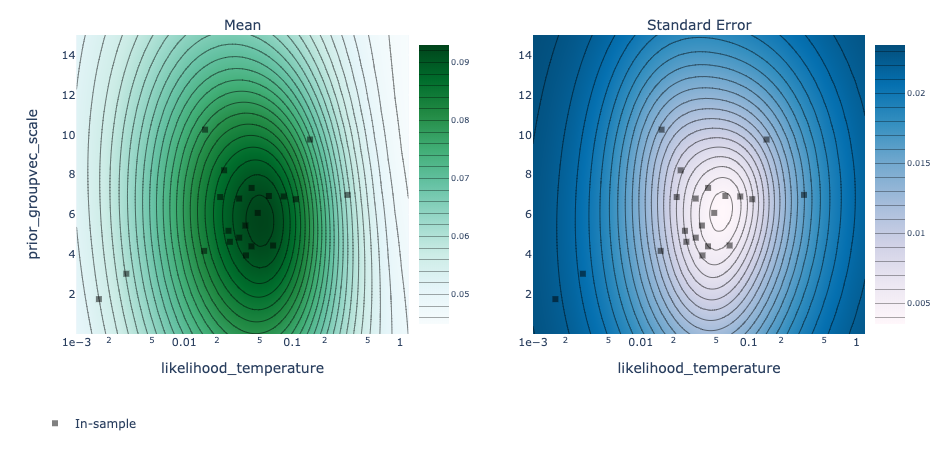}
\label{fig:contour_temp_kappa_mu}
\end{figure}

\begin{figure}
\caption{Bayesian Optimization: Marginal effects of $sigma_{max}$ vs. learning rate $l$}
\centering
\includegraphics[width=1.0\textwidth]{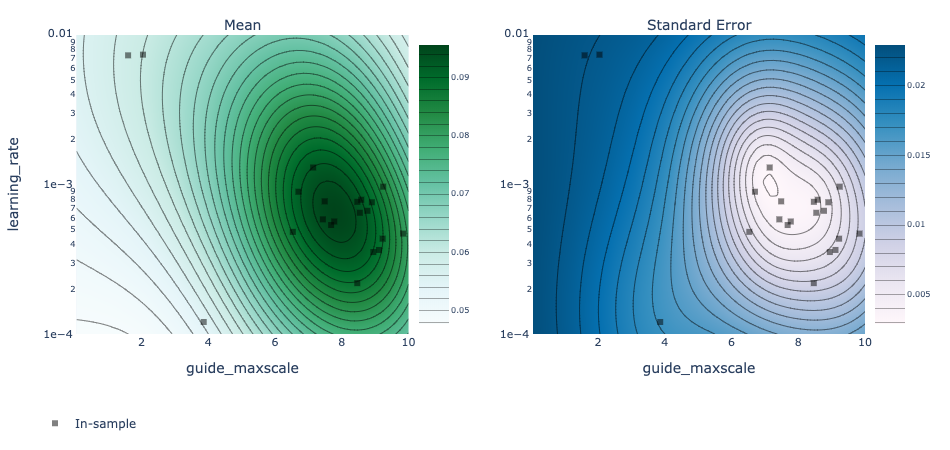}
\label{fig:contour_sigma_max_lr}
\end{figure}

For analysis, we also provide two surface plots on the marginal effects of different hyperparameters on the validation hitrate.
These figures are found in figure \ref{fig:contour_temp_kappa_mu} ad \ref{fig:contour_sigma_max_lr}.
We see that the Bayesian Optimization spends a proportionally larger time in the hyperparameter space that gives larger hitrates, as expected.

\section{ELBO optimization} \label{sec:optimization}
A good approximation of the posterior will be obtained by minimizing the  KL-divergence between the posterior and the variational distribution with respect to the variational parameters $\phi$.
This is equivalent to maximizing the Evidence Lower Bound (ELBO) over $\phi$ (see e.g. \cite{Blei2017}):
\begin{equation} \label{eq:elbo}
    \ELBO(\phi) = E_{q_{\phi}(\theta)} \left [ \log p(data|\theta) + \log p(\theta) -\log q_{\phi}(\theta) \right ]
\end{equation}
The expected approximate gradient of equation (\ref{eq:elbo}) with respect to the variational parameters $\phi$ is the following (see e.g. (\cite{Ranganath2014}) for derivation) :
\begin{equation} \label{eq:gradient_elbo}
    \nabla_{\phi} \ELBO(\phi) \approx \frac{1}{S} \sum_{s=1}^S \nabla_{\phi} \log q_{\phi}(\theta(s)) (\log p(\text{data subset}|\theta) + \log p(\theta)-\log q_{\phi}(\theta(s)) )
\end{equation}
where $\theta(s) \sim q_{\phi}(\theta)$ and  $S>0$ is the number of Monte Carlo samples needed to approximate the expectation and $p(\text{data subset}|\theta)$ is the likelihood of a minibatch of $\batchsize$ users, scaled by $\frac{1}{\batchsize}$.

Using stochastic optimization we can update the variational parameters sequentially 
\begin{equation} \label{eq:gradient_update}
    \phi_{b+1} = \phi_b + \lr \widehat{\nabla_{\phi} \ELBO(\phi_b)}
\end{equation}
where $\lr > 0$ is tunable learning rate parameter, $b>0$ is a batch iterator and $\widehat{\nabla_{\phi} \ELBO(\phi)}$ is an unbiased approximation by sampling both the Monte Carlo estimate and a subset of the data.

\subsection{Optimization settings}
It is not simple to optimize a large-scale variational inference problem. 
To ensure stable results, in our initial exploration of the problem we found that certain parameter settings gave better and more stable results.
The configurations can be found in Table \ref{tbl:optim_settings}.

\begin{table}[ht] 
\caption{
Various optimization settings we found to give stable and better results. 
}
\begin{tabular}{  l |  p{7cm} }
\hline\noalign{\smallskip}
\textbf{Action}      & \textbf{Description}   \\
\noalign{\smallskip}\hline\noalign{\smallskip}
    Batch size          & Larger batch sizes did generally seem to converge to better solutions. This could be due to reduced variance in estimation of (\ref{eq:gradient_elbo}) \\
    
    Stopping criteria   & We use a validation holdout maximum a posteriori likelihood estimation to determine stopping criteria. I.e. we would declare early stopping if the maximum a posteriori log likelihood of the validation data has not improved in 25 iterations of the training dataset. \\
    
    Scale parameter initialization & $\sigma_{\theta_k}$ set equal to some small positive number. When initialized to larger values then the posterior scales quickly increased and exploded to unreasonable large numbers. At the same time the maximum a posteriori estimate did not improve. We believe this effect can be attributed to very noisy gradient estimates in  (\ref{eq:gradient_elbo}). \\

\noalign{\smallskip} \hline
\end{tabular}
\label{tbl:optim_settings}
\end{table}

\end{document}